\documentclass[conference]{IEEEtran}
\IEEEoverridecommandlockouts
\usepackage{cite}
\usepackage{amsmath,amssymb,amsfonts}
\usepackage{algorithmic}
\usepackage{graphicx}
\usepackage{textcomp}
\usepackage{xcolor}
\usepackage{soul}
\usepackage{multirow}
\usepackage{tabu}                      
\usepackage{booktabs}                  
\usepackage{lipsum}                    
\usepackage{mwe}                       
\usepackage{mathptmx}                  
\def\BibTeX{{\rm B\kern-.05em{\sc i\kern-.025em b}\kern-.08em
    T\kern-.1667em\lower.7ex\hbox{E}\kern-.125emX}}
\begin{document}

\title{Using Motion Forecasting for Behavior-Based Virtual Reality (VR) Authentication}

\author{\IEEEauthorblockN{Mingjun Li, Natasha Kholgade Banerjee, Sean Banerjee}
\IEEEauthorblockA{Clarkson University, Potsdam, NY, USA \\
\texttt{\{mingli,nbanerje,sbanerje\}@clarkson.edu}}
}

\maketitle

\begin{abstract}
Task-based behavioral biometric authentication of users interacting in virtual reality (VR) environments enables seamless continuous authentication by using only the motion trajectories of the person’s body as a unique signature. Deep learning-based approaches for behavioral biometrics show high accuracy when using complete or near complete portions of the user trajectory, but show lower performance when using smaller segments from the start of the task. Thus, any system designed with existing techniques are vulnerable while waiting for future segments of motion trajectories to become available. In this work, we present the first approach that forecasts future user behavior using Transformer-based forecasting and using the forecasted trajectory to perform user authentication. Our work leverages the notion that given the current trajectory of a user in a task-based environment we can forecast the future trajectory of the user as they are unlikely to dramatically shift their behavior since it would preclude the user from successfully completing their task goal. Using the publicly available 41-subject ball throwing dataset of Miller et al. we show improvement in user authentication when using forecasted data. When compared to no forecasting, our approach reduces the authentication equal error rate (EER) by an average of 23.85\% and a maximum reduction of 36.14\%. 
\end{abstract}

\begin{IEEEkeywords}
VR biometrics, Transformers, Motion forecasting
\end{IEEEkeywords}

\section{Introduction}

VR has seen rapid growth in critical domains such as education~\cite{noah2021exploring,agbo2021application}, nursing and medicine~\cite{hamilton2021immersive,shorey2021use,barteit2021augmented,clarke2021virtual}, retail~\cite{pizzi2019virtual,xue2019virtual}, personal finance~\cite{campbell2019uses,weise2016virtual}, and healthcare~\cite{munoz2022immersive,mehrabi2022immersive,karaosmanoglu2022canoe}. As VR devices become more affordable and portable, it is likely that more users will adopt them for everyday use. As a result, such critical applications must contain mechanisms to identify or authenticate a user. Early research in securing VR systems adopted traditional PIN and password-based credentials~\cite{alsulaiman2008three,alsulaiman2006novel,gurary2017leveraging,george2020gazeroomlock,yu2016exploration,george2017seamless,olade2020exploring,funk2019lookunlock,george2019investigating}. Techniques based on a password or a PIN are known to be unsafe, as once the malicious agent gains access to the credentials, the user's account is immediately compromised. The malicious agent may be an external agent or the genuine user deliberately handing their credentials to an ally to defeat a system. A genuine user handing over credentials to an ally is a problem in environments where cheating or non-adherence is a prevalent issue, such as education or healthcare. 

Recently, a large body of work has emerged to use user behavior in VR as a biometric signature for securing access~\cite{mustafa2018unsure,mmm2019vr,pfeuffer2019behavioural,ajit2019combining,miller2019,mathis2020knowledge,mathis2020rubikauth,mathis2020fast,miller2020within,olade2020biomove,miller2020personal,miller2021using,liebers2021understanding}. Identification accuracies have reached upwards of 95\%~\cite{mathis2020knowledge,miller2020within,olade2020biomove,miller2020personal,miller2021using}, and these approaches investigate identification and authentication for a number of tasks, e.g., watching a video, throwing a ball, turning a cube, and making a golf swing where tasks are easily remembered and largely repeatable. A fundamental limitation of existing work on behavior-based biometrics for securing VR systems is the reliance on complete or near complete trajectories of user behavior. Kupin et al.~\cite{mmm2019vr}, Ajit et al.~\cite{ajit2019combining}, and Miller et al.~\cite{miller2020within,miller2021using} demonstrate that using smaller portions of the entire trajectory yields lower performance, with large performance drops when less than 80\% of the trajectory is used.

\begin{figure}
\includegraphics[width=\linewidth]{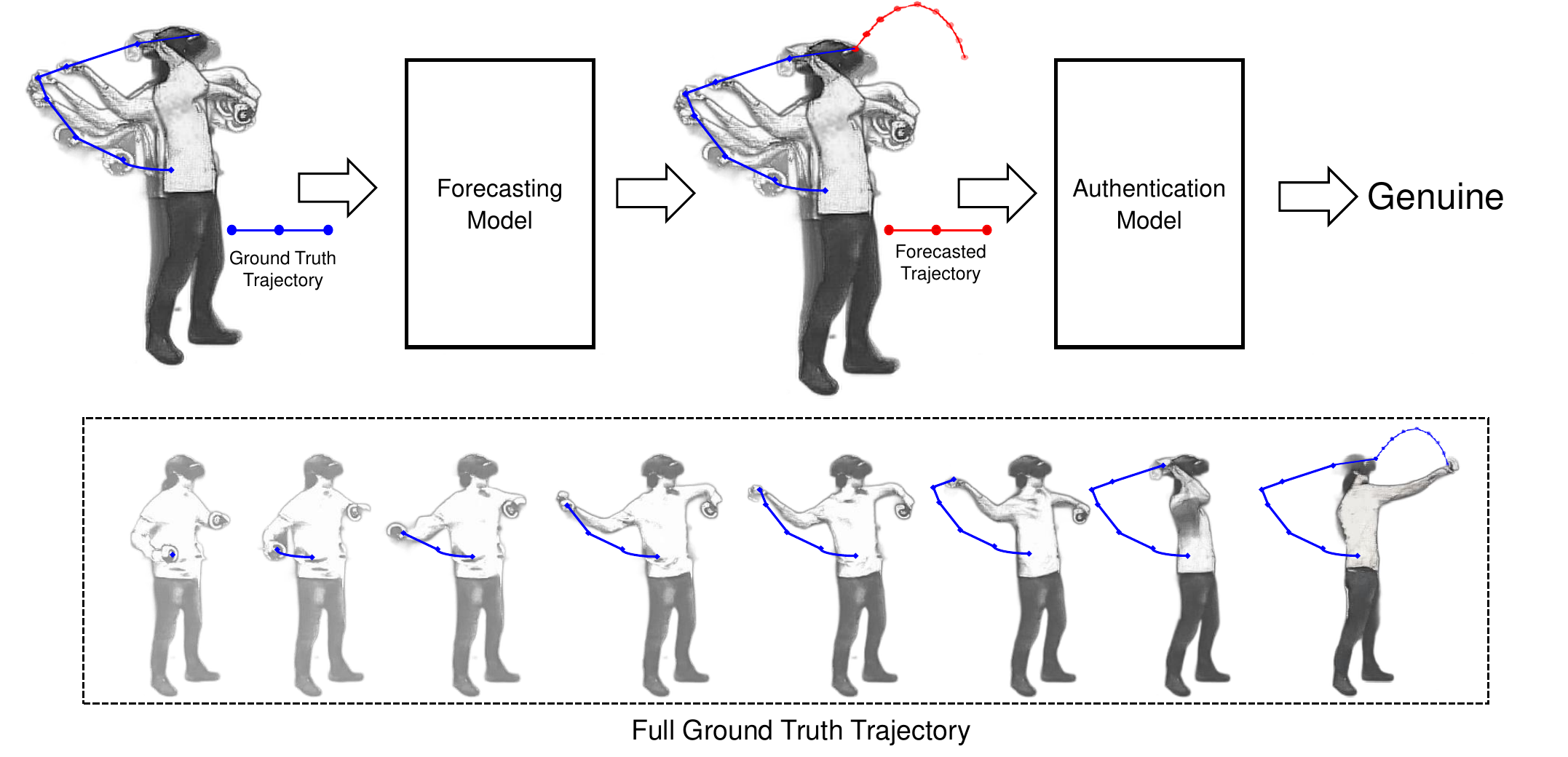}
\caption{In our approach, we utilize the ground truth input trajectory to forecast the future trajectory, which is subsequently merged with the input trajectory to authenticate users. When compared to no forecasting, our approach reduces the authentication equal error rate (EER) by an average of 23.85\% and a maximum reduction of 36.14\%. The upper portion of the figure outlines our approach, while the lower portion shows the complete ground truth trajectory.}
\end{figure}

In this paper, we propose the \textbf{first approach that uses motion forecasting to predict plausible future motion trajectories.} Using motion forecasting for path, or trajectory, planning has received increased attention due to the growth of autonomous driving systems where motions of objects must be forecasted ahead of time~\cite{zhou2022hivt,huang2022multi,kong2022human,yuan2021agentformer}. We train a Transformer-based model~\cite{vaswani2017attention,devlin2018bert,zhou2021informer} to forecast the user's motion behavior trajectory for a period of time in the future using a portion of the starting trajectory. During authentication, our approach uses the past user behavior and combines it with the forecasted trajectories. In our approach, we use the predicted motion trajectories to perform authentication and demonstrate that we can achieve higher accuracies. Using the 41-subject ball-throwing dataset of Miller et al.~\cite{miller2022temporal,miller2022combining} for testing, we show in Section~\ref{sec:results} that we consistently achieve lower equal error rate (EER, the standard metric for evaluating biometric systems~\cite{jain2007handbook}) with forecasting than without for all window sizes, with a maximum drop of 0.039 in EER from without forecasting to with forecasting. With no forecasting, our best EER using FCN as the classifier is 0.062 for a window size of 75. With forecasting, we can reduce the window size to as low as 45 and obtain a lower EER (0.061) by forecasting 40 future timestamps. Our overall lowest EER using FCN as the classifier is 0.052 and is obtained at a window size of 65 and forecasting 30 timestamps. When looking at the Transformer encoder as the classifier, our best EER without forecasting is 0.057 at window size 75. With forecasting, our window size can be as low as 45 and yield a lower EER (0.053) by forecasting 50 timestamps. The overall lowest EER we obtain with the Transformer encoder as the classifier is 0.048 for window size 65 and forecasting 30 timestamps. Our code can be downloaded at: \texttt{http://tinyurl.com/forecastauth}.

\section{Related Work}
A growing number of approaches have arisen in the last decade on VR authentication. Their impact is supported by recent literature survey~\cite{jones2021literature,giaretta2022security}, a Systematization of Knowledge (SoK)~\cite{stephenson2022sok}, and position papers~\cite{alt2022beyond} making recommendations on the future of VR security, e.g., integration of multiple modalities such as physiological (e.g., face) and behavioral biometrics~\cite{alt2022beyond}, and the need to enable cross-device or cross-context security~\cite{alt2022beyond}. 

\paragraph{Passwords and PINs} Traditional work in providing security in VR environments has largely addressed the question of enabling users to enter credentials such as passwords in the VR environment. The focus of investigation for these approaches tends to be to provide resistance to shoulder-surfing attacks, and to ensure usability by assessing how convenient it is for the user to enter the password. Some approaches focus on directly translating the concept of a 2D password to the VR environment. Mechanisms to seek entry of alphanumeric passwords can be challenging, as using controllers or gaze to interact with a VR keyboard can be cumbersome. As such, 2D passwords tend to largely be lock patterns similar to those on smart devices. Studies have investigated the security and usability of lock patterns imposed on axis aligned or inclined planes~\cite{yu2016exploration,george2017seamless}. The studies have conducted evaluations of the type of interaction that is most convenient for usability, e.g., pointing and pulling the controller trigger versus using a VR stylus or clicking the trackpad~\cite{george2017seamless}. Evaluations have also been conducted of resistance of VR lock patterns to shoulder surfing~\cite{olade2020exploring}. Other approaches advocate the use of the 3D space to provide novel 3D passwords. These passwords may either consist of a unique selection of 3D virtual objects~\cite{yu2016exploration,george2019investigating,funk2019lookunlock,george2020gazeroomlock}, or of a unique sequence of actions performed by the user in the virtual environment~\cite{gurary2017leveraging}. Inspiration for the latter comes from analyses of the action space for 3D passwords in a graphical environment and the ability of the action space to provide security guarantees~\cite{alsulaiman2006novel,alsulaiman2008three}. 3D passwords based on virtual object selection may be entered by selecting the object permutation using a controller~\cite{gurary2017leveraging}, using gaze to point at the objects comprising the sequence~\cite{funk2019lookunlock}, or using a combination of gaze- and controller-based selection~\cite{george2020gazeroomlock}. 

Most studies demonstrate high shoulder-surfing resistance of password entry mechanisms, with 3D passwords being more resistant to 2D passwords~\cite{yu2016exploration}. However, if an attacker gains access via an alternate mechanism, e.g., through a man-in-the-middle attack, the system is immediately compromised. Additionally, while 3D passwords may provide higher security guarantees~\cite{yu2016exploration}, since they are an uncommon form of password entry, users may face lower usability if memorizing the 3D password is more challenging or requires more time than traditional credentials. Gurary et al.~\cite{gurary2017leveraging} demonstrate that retention of 3D passwords based on action sequences is significantly higher than 2D passwords. George et al.~\cite{george2020gazeroomlock} show that multimodal approaches that combine gaze with controller-based selection reduce error rate in password entry, indicating higher memorability over unimodal approaches. Usability of a password entry mechanism depends on how familiar users are with the VR system and how comfortable they are in performing the interaction. Yu et al.~\cite{yu2016exploration} demonstrate that users found entering simple combinations of 3D passwords using the LeapMotion to be less usable than entering 2D passwords. George et al.~\cite{george2020gazeroomlock} demonstrate that using gaze in conjunction with controller selection provides the highest usability. However, more studies are needed to evaluate how users perceive usability and memorability during long-term use. Any form of password entry hampers continuous authentication, as it requires users to stop their activity to enter credentials. Long credential-entry times could prove detrimental to performance during, for instance, a high-stress examination or military routine, or hazardous to an operation during VR-based remote teleoperation.

\paragraph{Behavioral Biometrics} Given the challenges with traditional credentials and the lack of biometric scanners embedded in VR devices, a large body of work has emerged on leveraging user behavior in VR as a biometric. Currently, user VR behavior is largely modeled by tracking the motions of the headset, hand controllers, and objects in the VR space while the user performs interactions in the VR environment. Mustafa et al.~\cite{mustafa2018unsure} provide an approach that uses support vector machines to classify users based on head movement while users listen to music on a Google Cardboard. Kupin et al.~\cite{mmm2019vr} use nearest neighbors to automatically identify users from the trajectories of the dominant hand controller as users throw a ball at a target in VR. To garner maximum benefit from the comprehensive motion of the user in the environment, most current behavioral biometrics research leverages a multimodal approach that combines features from motion tracks of the headset and controllers. Ajit et al.~\cite{ajit2019combining} use a perceptron to classify distances from position and orientation features acquired from the headset and hand controller trajectories in the input and library sessions for a user performing the ball-throwing action of Kupin et al.~\cite{mmm2019vr}. Miller et al.~\cite{miller2020within} extend the method of Ajit et al. to include velocity, angular velocity, and trigger features for performing identification using ball-throwing sessions provided within a single VR system, and using sessions spanning multiple VR systems. Pfeuffer et al.~\cite{pfeuffer2019behavioural} evaluate random forests and SVMs on aggregate statistics drawn from unary features and pairwise relationships established amongst the headset, controllers, and target VR objects for activities such as picking, pointing, and grabbing. 

Miller et al.~\cite{miller2020personal} evaluate multiple learning algorithms on a dataset of users watching 5 videos and performing question answering on the videos. Olade et al.~\cite{olade2020biomove} investigate nearest neighbors and support vector machines for classifying users performing dropping, grabbing, and rotating from their motion trajectories. To improve accuracy while removing reliance on hand-crafted features, more recent approaches have navigated toward using deep learning. Mathis et al.~\cite{mathis2020knowledge} use 1D convolutional neural networks (CNNs) to classify sliding window trajectory snippets from the headset and hand controllers for users using pointing interactions to select passwords on a cube. Liebers et al.~\cite{liebers2021understanding} use recurrent neural networks to classify users performing bowling and archery activities in VR. Miller et al.~\cite{miller2021using} use Siamese networks to learn cross-system relationships for improving identification and authentication when library and input data spans multiple VR systems.  

The reliability of VR behavioral biometrics depends on the consistency of user behavior in VR. Several VR datasets~\cite{olade2020biomove,mathis2020knowledge,miller2020personal} typically involve users providing data within a single session over the span of a few minutes, where behavior variability may be limited. Work on the temporal effect on behavioral biometrics has explored the impact of short-, medium-, and long-timescale user behavior variations~\cite{miller2022temporal} and reveals two concerns: (1)~authentication performance degrades when system-specific noise increases~\cite{miller2021using,miller2022combining}, and (2)~authentication improvement requires training with data from varying temporal separations. Behaviors explored in VR thus far are repeatable actions with clear spatial extents such as throwing a ball, bowling, or shooting an arrow, or action primitives such as picking or pointing. The approaches explored so far may be implementable for complex activities such as physical therapy or military drills that have necessarily repeatable routines. Our work leverages the repeatable nature of tasks in VR to forecast future user behavior based on the past behavior. Our work has a significant advantage in requiring only the initial motion behavior as the forecasted behavior can be leveraged during authentication. Thus, unlike existing work, our work enables authentication with lesser data which limits the amount of time the system is vulnerable.  

\section{Dataset}

\begin{figure}[t!]
    \centering
    \includegraphics[width=\linewidth]{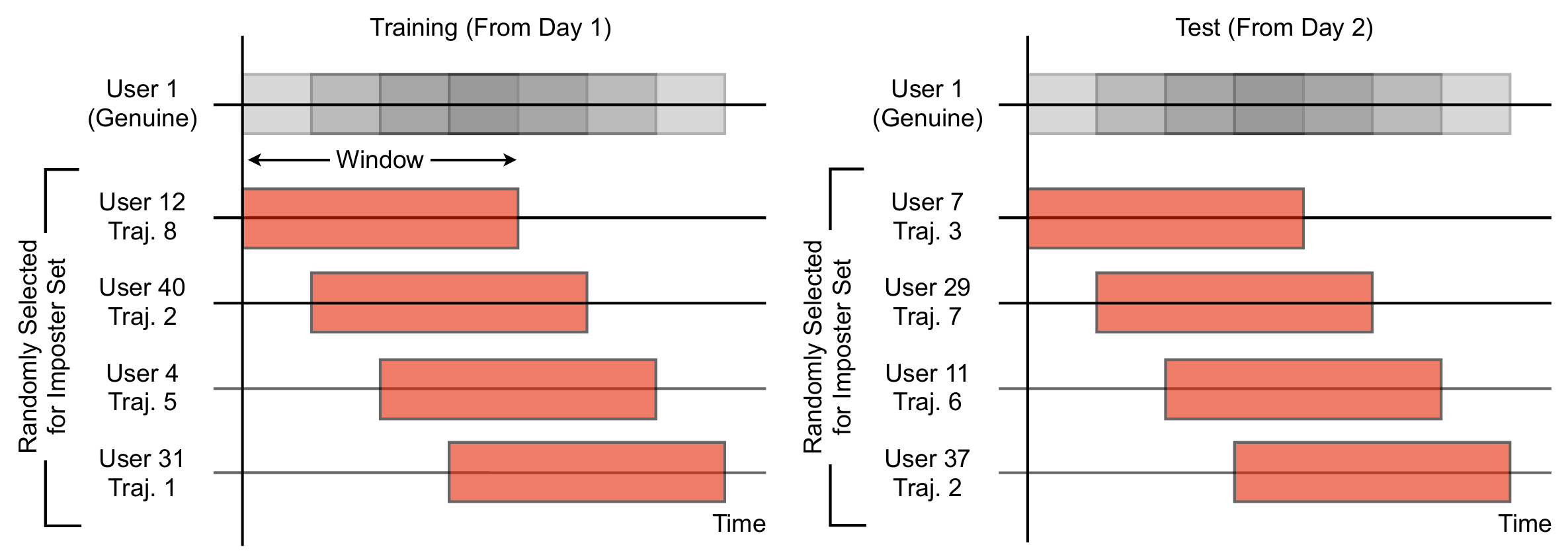}
    \caption{Left: To create the training set for authentication, we evenly sample sliding windows of size $n$ from day 1 trajectories of the genuine user. To create the impostor set, for each genuine sliding window, we randomly sample a subject and day 1 trajectory from the remaining users, and select a window from the trajectory sample at the same temporal location as the genuine sliding window. Right: we repeat the process with day 2 trajectories to create the test set, ensuring that the random ordering of subjects/sessions is different. }
    \label{fig:datageneration}
\end{figure}

\begin{figure*}[t!]
    \centering
    \includegraphics[width=\linewidth]{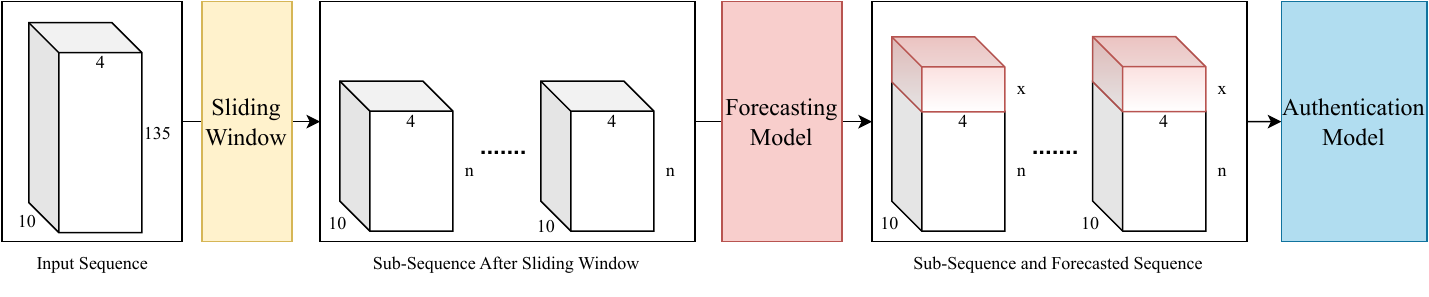}
    \caption{Pipeline flowchart of our proposed approach. In the first step, the input data is processed using the sliding window technique to generate sub-sequences. These sub-sequences are then fed into the forecasting model, which generates the forecasted sequence. The forecasted sequence is then concatenated with the original input data to form a combined sequence. Finally, the combined sequence is fed into the classifier for authentication. $135$, $10$, and $4$  represent the total timestamps in raw data, number of sessions, and number of features for each session, respectively.}
    \label{fig:pipeline}
\end{figure*}
 
We use the dataset of Miller et al.~\cite{miller2020within, miller2021using} consisting of 41 right-handed subjects performing a ball-throwing task using 3 VR systems as it is publicly available. Approximately 10\% of the population is left-handed~\cite{papadatou2020human} making it challenging to obtain sufficient samples. The task consists of a user picking up a ball on a pedestal and throwing it at a target directly in front of them. Users provide data using an HTC Vive, HTC Vive Cosmos, and Oculus Quest across two days separated by at least 24 hours. On each day users provide 10 sessions, for a total of 20 sessions per VR system. The physical characteristics and locations of the ball, target, and pedestal remain constant throughout the procedure across each trial and session. The dataset consists of $x$, $y$, and $z$ position and orientation values as Euler angle rotations around $x$, $y$, and $z$ axes for the headset and hand controllers, as well as trigger pressure for the controllers. The trigger pressure represents the amount of force applied to the trigger on the controller. For this paper, we only use data from the HTC consisting of the right-hand controller trajectory position and trigger pressure. 

\subsection{Data Preparation}

We extract data over each session for each subject by sliding a window over the session data. We denote a session as $s^{u}_{i}$, where $u$ refers to the user id, and $i$ refers to the session number. Each session $s^{u}_{i}$ is a matrix of real numbers of size $T \times f$, where $T$ refers to the total number of timestamps and $f$ refers to the number of features. For each session we apply a sliding window of size $n \times f$ and stride $l$ to $s^{u}_{i}$ along the temporal dimension to extract time-varying chunks of the session data. 

\subsection{Impostor Data Generation}

Each session in the dataset utilized for this study represents authentic data from the subjects under investigation. However, to enable the network to learn effective identification and authentication capabilities, it is necessary to incorporate impostor data into the training process. Rather than generating arbitrary data, we obtain impostor data by extracting from other users selected at random, and each piece of impostor data has the same start and end point of time as that in the corresponding genuine data, as shown in Figure~\ref{fig:datageneration}. The random selection allows us to diversely represent the patterns and behaviors of an actual adversary, while still being independent from the genuine data of the current user. To ensure a fair comparison between the genuine and impostor data, we start the impostor data at the same timestamp as the genuine data for the current user, and make sure that the length of the impostor data matches that of the genuine data, so that the two types of data have the same temporal alignment. Using this approach to extract impostor data, we build a more realistic and balanced dataset for neural network training.

\begin{figure}[tbp]
    \centering
    \includegraphics[width=.45\textwidth]{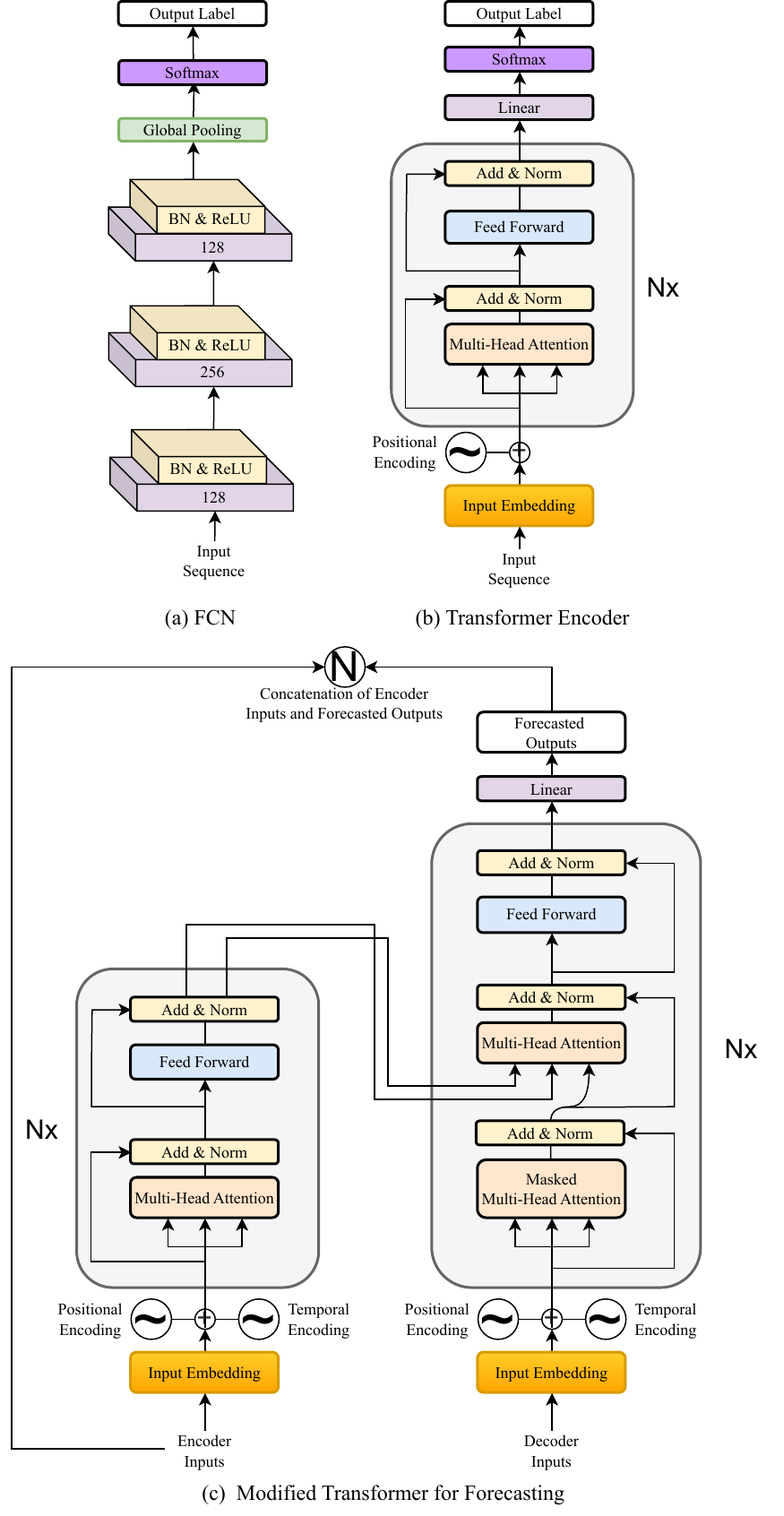}
    \caption{(a) an FCN and (b) a Transformer Encoder as for authentication. (c) We use a modified Transformer for forecasting.}
    \label{fig:models}
\end{figure}

\section{Motion Forecasting}

As shown in Figure~\ref{fig:pipeline}, our method involves breaking down the input time series data into segments, with each segment containing a fixed number of timestamps. For each segment, we train a model based on the Informer~\cite{zhou2021informer}, as shown in Figure~\ref{fig:models}(c), to forecast the subsequent time behavior trajectory. When forecasting, we avoid making multiple calls to the Transformer as it causes errors to accumulate, as each next-timestep forecast will depend on the prior. Thus, we generate the entire forecasted trajectory at once as it avoids error accumulation. The forecasted output is then combined with the real input data, resulting in semi-synthetic complete data. The concatenated data is then input into a classifier as shown in Figure~\ref{fig:models}(a) and Figure~\ref{fig:models}(b) for authentication. 

\paragraph{Feature Representation}
We use learned embeddings that map each timestamp's data to a higher-dimensional space of size $d_{model}$, to extract information from the input data,  which is originally in a 4-dimensional space ($x$, $y$, $z$ coordinates and trigger pressure measurement). We use the same approach as Vaswani et al.~\cite{vaswani2017attention} to preserve positional information of the input sequence. We encode the position information of each timestamp data using sine and cosine functions
\begin{align}\label{pos1}
    PE(t, 2i) &= \sin\left( {t} / \left({10000^{2i / d_{model}}}\right)\right)~\textrm{and}\\
    PE(t, 2i+1) &= \cos\left({t} / \left({10000^{2i / d_{model}}}\right)\right),
\end{align}
where $t$ is the timestamp and $i$ is the dimension. With this positional encoding, our model learns to distinguish and relate temporal information based on their positions.
Using the approach of Zhou et al.~\cite{zhou2021informer}, which encodes long-range time attributes such as year, month, week, and day to scalars, we define the function $TE(t)$ as
\begin{equation}\label{pos3}
    TE(t) = t / T - 0.5,
\end{equation}
to encode the short-range time data represented in milliseconds to a scalar in the range of -0.5 to 0.5. The value $t$ represents the timestamp and $T$ is the total number of timestamps. 
The value $d_{model}$ also represents the dimension of the output of positional and temporal encoding. We add the learned input embeddings, positional encodings, and time encodings, enabling us to represent the input data in a high-dimensional space that preserves positional and temporal relationships between timestamps.

\paragraph{Encoder} 
Our encoder consists of multiple encoder layers, where each encoder layer is composed of a multi-head attention sub-layer, a position-wise fully connected feed-forward sub-layer, residual connection operation~\cite{he2016deep}, and layer normalization~\cite{ba2016layer} as shown in Figure~\ref{fig:models}(c). The multi-head attention sub-layer enables parallel computations in $n_{head}$ scaled single-head dot product self-attentions, with each self-attention focusing on different parts of the input sequence. The multi-head attention allows our model to capture more complex relationships between the input elements. As defined in the original Transformer paper~\cite{vaswani2017attention}, each single-head dot product attention unit computes a weighted sum of the values $V$ of the input sequence, as 
\begin{equation}\label{attn}
    Attention(Q, K, V) = \textrm{softmax}\left(\frac{QK^T}{\sqrt{d_K}}\right)V,
\end{equation}
where the weights are determined by the similarity of the query vector $Q$ and the key vector $K$ of each element, which is then scaled by the square root of the dimensionality of the key vector $d_K$ to ensure that the attention scores are not too large. We apply a softmax function to obtain a probability distribution over the weighted sum.
The residual connection operation~\cite{he2016deep} adds the output of the multi-head attention sub-layer to the original input to smooth the gradient flow during training and to facilitate learning of deeper representations. The position-wise dense feed-forward sub-layer applies a fully connected neural network to each element of the sequence independently. The fully connected sub-layer has an input and output dimension of $d_{model}$, and a hidden layer of dimension $d_{hidden}$. We perform layer normalization~\cite{ba2016layer} after each residual connection. In this work, we employ a stack of two identical encoder layers.

\paragraph{Decoder}

\begin{figure}[t]
   \centering
   \includegraphics[width=\linewidth]{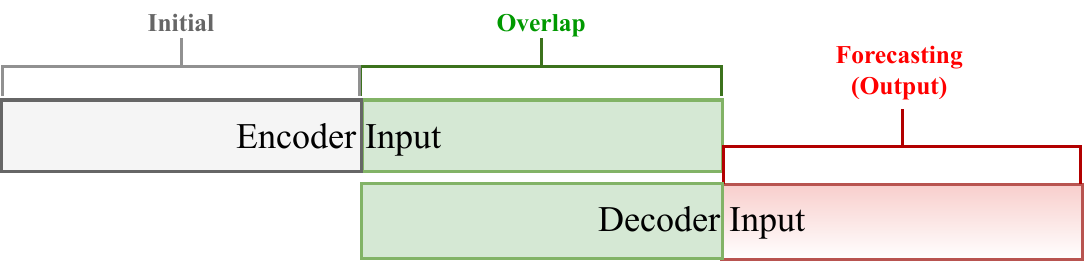}
   \caption{The input to the Encoder consists of the initial sequence (in gray) and the overlap sequence (in green), and the Decoder input consists of the overlap sequence (in green) and the sequence to be forecasted initialized with zeros (in red). }
    \label{fig:inout}
\end{figure}

We extract a subset with length $l_{overlap}$ from the input sequence of the encoder, as shown in Figure~\ref{fig:inout} in green. We initialize the region to be predicted, shown in red in Figure~\ref{fig:inout}, with zeros. We concatenate the encoder subset in green with the initialization in red to form the input to the decoder. The decoder inherits the learned patterns from the encoder. We apply input embedding, positional encoding, and temporal encoding to the decoder input to convert the input to a higher dimensional space. Similar to the traditional transformer decoder, we incorporate a masked self-attention sub-layer to correlate each element in the decoder input sequence and a masked cross-attention sub-layer to correlate the decoder input with the encoder output. The standard Transformer decoder~\cite{vaswani2017attention} operates on a one-step prediction basis, outputting the prediction result element by element. Their approach is not suitable for our goal of generating forecasted results of multiple future timestamps at once. To address this issue, we use a fully connected feed-forward sub-layer at the end of the decoder, so that our model outputs forecasting results of an arbitrary length of timestamps, $l_{forecasting}$, at a time. Similar to the encoder, we perform residual connection~\cite{he2016deep} and layer normalization~\cite{ba2016layer} after each sub-layer.

\setlength{\tabcolsep}{4.5pt}
\begin{table*}[tb]
  \caption{Equal Error Rate of No Forecasting. The abbreviation `WS' refers to the window size and the subsequent numbers in the same row denote the values of window size, and the last column is the average value of each row. `FCN' stands for Fully Convolutional Networks~\cite{wang2017time}, `TF' represents the Transformer encoder~\cite{vaswani2017attention}, `EER' represents the equal error rate (where lower values are preferable). Each row is the average of all $41$ subjects under the corresponding column.}
  \label{tab:exp1}
  \scriptsize%
  \small
  \centering%
  \begin{tabu}{c|ccccccccccccccc | c
  	}
  	\toprule
  	WS & 25 & 30 & 35 & 40 & 45 & 50 & 55 & 60 & 65 & 70 & 75 & 80 & 85 & 90 & 95 & Mean\\
  	\hline \hline
        
    FCN~\cite{wang2017time} & 0.121 & 0.109 & 0.101 & 0.091 & 0.082 & 0.080 & 0.082 & 0.061 & 0.075 & 0.061 & 0.062 & 0.064 & 0.071 & \textbf{0.048} & 0.066 & 0.078\\

        TF~\cite{vaswani2017attention} & 0.115 & 0.104 & 0.097 & 0.089 & 0.083 & 0.077 & 0.072 & 0.065 & 0.064 & 0.058 & \textbf{0.057} & 0.063 & 0.064 & 0.062 & 0.061 & \textbf{0.075}\\
  	\bottomrule
  \end{tabu}%
\end{table*}

\section{Authentication}
We compare two models for authentication as shown in Figure~\ref{fig:models}(a) and Figure~\ref{fig:models}(b), namely a Fully Convolutional Network (FCN)~\cite{wang2017time} and a Transformer encoder~\cite{vaswani2017attention}. We train one FCN/Transformer encoder per user. We obtain the genuine data from each user using the sliding window technique by extracting windows of window size $n$ and number of features $f$. The features represent the $x$, $y$, and $z$ positions of the right-hand controller and trigger pressure. Each window is of dimensions $n \times f$. We randomly select impostor data from the remaining subjects and each piece of the impostor data consists of the same timestamp data as the genuine data, as the starting point for all trajectories in the Miller dataset occurs when the user picks the ball off the pedestal by pulling the controller trigger. Randomly sampling multiple users enables covering a diverse range of speeds of performance in the impostor set. We evaluate the performance of the trained models on a set of previously unseen data after each training epoch. 

\paragraph{Fully Convolutional Network (FCN)} 
We use the FCN architecture in Wang et al.~\cite{wang2017time} which consists of three convolutional blocks, each with a convolutional layer and a 1D kernel. To enhance convergence and improve generalization, batch normalization layers~\cite{ioffe2015batch} are applied after each convolutional layer, followed by ReLU activation layers at the end of each block. A GAP layer~\cite{lin2013network} is employed after these three blocks, and a softmax layer provides the final output as shown in Figure~\ref{fig:models}(a). Mathis et al.~\cite{mathis2020knowledge} show that the FCN outperforms other approaches for VR security. 

\paragraph{Transformer Encoder}
Though FCNs have shown success in time series classification, they lack strength of attention networks in relating different portions of the trajectories. To capture intra-trajectory relationships in authentication, we evaluate a second network that uses the encoder of the Transformer architecture~\cite{vaswani2017attention} to perform authentication as shown in Figure~\ref{fig:models}(b). The Transformer encoder has the ability to capture global correlations between each element in an input sequence by the multi-head self-attention mechanism~\cite{vaswani2017attention}, which is an important characteristic for analyzing time series data. We employ the encoder only owing to its ability to extract meaningful features from the input sequence rather than generating a list of output elements. We eliminate the temporal encoding used for the forecasting Transformer. We eliminate temporal encoding as for the second (authentication) step, we use the Transformer encoder for a simpler task, i.e., binary classification of genuine vs impostor. The task of forecasting in the first step benefits from explicit temporal dependence~\cite{zhou2021informer} to model time series progression. With binary classification, removing temporal encoding and retaining positional encoding reduces compute time with minimal impact on results.

\paragraph{Loss Functions}
During training, we optimize for the model parameters by minimizing the loss 
\begin{align}\label{LOSS}
    \mathtt{L} = \mathtt{L}_{L} + \lambda_{F}  \mathtt{L}_{F} + \lambda_{T}  \mathtt{L}_{T}.    
\end{align}
In Equation~\eqref{LOSS}, $\mathtt{L}_{F}$, represented as
\begin{equation}\label{LF}
    \mathtt{L}_{F} = (1/|W|) \Sigma_{w \in W} MSE(Tra_{pred}, Tra_{gt}).
\end{equation}
measures the discrepancy between the forecasted right-hand controller trajectory and the corresponding ground truth trajectory. In Equation~\eqref{LF}, $MSE$ represents the mean squared error loss function, $Tra_{pred}$ and $Tra_{gt}$ are the forecasted trajectory and ground truth trajectory respectively. $|W|$ denotes the total number of windows while $w$ stands for a particular window of the whole window set $W$. We define
\begin{equation}
     \mathtt{L}_{T} = (1/|w|) \Sigma_{t \in w} BCE(Tri_{pred}, Tri_{gt}), \textrm{ and}
     \label{eq:bcetri}
\end{equation}
\begin{equation}
    \mathtt{L}_{L} = (1/|W|) \Sigma_{w \in W} BCE(Label_{pred}, Label_{gt}),
    \label{eq:bcelab}
\end{equation}
where BCE is the binary cross-entropy loss function. Equation~\eqref{eq:bcetri} provides BCE for trigger pressure, $Tri$, and Equation~\eqref{eq:bcelab} for forecasted authentication label, $Label$. We set the value of the ground truth label to 1 for a genuine user and 0 for an impostor. The value $t$ refers to a specific timestamp in the window $w$, and subscripts $pred$ and $gt$ stand for generated outputs and ground truth. We use the notation $\lambda_{F}$ and $\lambda_{T}$ in Equation~\eqref{LOSS} to denote the weights for the loss terms $\mathtt{L}_{F}$ and $\mathtt{L}_{T}$. We use Adam~\cite{kingma2014adam} as the optimizer. 

\paragraph{Implementation Details}

We conducted training using a 12-core Ryzen 9 5900X 3.7 GHz CPU with an NVIDIA GeForce RTX 4090 GPU. Training was conducted over 200 epochs for all models. Training times range over 80-151 sec for FCN and 110-218 sec for the Transformer.

\section{Experimental Results}
\label{sec:results}
We use the day 1 data of $41$ subjects in the Miller et al.~\cite{miller2020within,miller2021using} dataset for training the network, and day 2 data for evaluating the network's performance. In our `No Forecasting Experiment', we train the FCN and Transformer encoder to predict the classification label of the input data directly. In `Authentication with Forecasting Experiment', we use our proposed approach to forecast trajectory data and then combine it with the input data before performing classification. We evaluate our approach by computing the \textbf{equal error rate (EER)}. The EER indicates the point at which the false acceptance rate is equal to the false rejection rate, the lower the EER value, the better performance of the model. 

\subsection{No Forecasting Experiment}
We vary the size of the sliding window, $l_{window}$, from $25$ to $95$ with a step size of $5$. In this experiment, we only compute the BCE loss using Equation~\ref{eq:bcelab}. For the FCN, we use three convolutional blocks, each of them contains a convolutional layer with a filter size of \{$128$, $256$, $128$\} and a 1D kernel size of \{$8$, $5$, $3$\}, respectively. We use Adam~\cite{kingma2014adam} as the optimizer with a learning rate of $0.001$. For the Transformer, we perform input embedding and positional encoding to the input sequence that projects the input data from its original dimension to $d_{model} = 512$. We employ a stack of two encoder layers, which are identical in structure, to process the input data in the classification task. Each encoder layer has a $n_{head} = 8$ multi-head attention sub-layer in it. Lengths of query, key, and value vectors for all the heads are $d_q = d_k = d_v = 64$. We use the Adam~\cite{kingma2014adam} optimizer with a learning rate of $0.0001$.

Table~\ref{tab:exp1} summarizes the results of the No Forecasting Experiment, where the abbreviation `WS' in the first line refers to window size, and the subsequent numbers denote the specific values of window size we employed. The acronyms used in this table are as follows: `FCN' stands for Fully Convolutional Network~\cite{wang2017time}, `TF' represents the Transformer encoder~\cite{vaswani2017attention}, and `EER' represents the equal error rate (where lower values are preferable). Each row of Table~\ref{tab:exp1} represents the average testing EER of all $41$ subjects under the corresponding column.

Values from Table~\ref{tab:exp1} reveal that the EER of the two models exhibits a similar trend, decreasing with an increase in window size. We observe that the overall performance of TF is better than that of FCN. For most window sizes, TF provides lower EER values, except for window sizes $45$, $60$, and $90$. However, the lowest EER among all window sizes is achieved by FCN at window size $90$. We also find that for each of the models, FCN performs best when WS $=90$, whereas the Transformer encoder performs best when WS $=75$. We conclude from the last column in Table~\ref{tab:exp1} that the Transformer encoder outperforms the FCN model. It also demonstrates that the performance of the models is influenced by the window size, indicating that the choice of window size plays a crucial role in determining the effectiveness of the models.

\subsection{Authentication with Forecasting Experiment}
We aim to generate a forecasted sequence of data with a length of $l_{forecasting}$ based upon data within a window of length $l_{window}$, where $l_{window}$ is determined as the sum of the initial length $l_{initial}$ and the length of the overlapping data $l_{overlap}$ as shown in Figure~\ref{fig:inout}. We investigate various combinations of $l_{window}$ and $l_{forecasting}$. We vary $l_{window}$ from $25$ to $85$ at a step size of $10$, and $l_{forecasting}$ from $10$ to $70$ with a step size of $10$. We choose to terminate the sliding window process at a window size of 85, as 85 exceeds more than half of the original data length of 135 timestamps, and our goal is to evaluate the performance of using a reduced subset of data for authentication. We conduct multiple trials with varying $l_{overlap}$ sizes, from $5$ to $l_{window} - 5$ with stride $5$ for each set of fixed $l_{window}$ and $l_{forecasting}$ pairs, to investigate whether the length of the overlap area,  $l_{overlap}$, has an impact on the accuracy of the forecasted trajectory. For the Transformer-based forecasting model, we use $3$ encoder layers and $1$ decoder layer. The dimension of this model is $d_{model} = 512$, with a total of $n_{head} = 8$ attention heads for each layer. The query, key, and value dimensions are set to $d_q = d_k = d_v = 64$. We use a fully connected layer with dimension $d_{hidden} = 2048$. We use the Adam~\cite{kingma2014adam} optimizer with a learning rate of $0.0001$.

\begin{table}[t!]
  \caption{Forecasting Trajectories MSE Scores. `WS' is window size, and `+x' refers to the length of forecasted sequence is x.}
  \label{tab:fore_mes}
  
  \small
  \centering%
  \begin{tabu}{
             c| *{8}{c}
            *{9}{c}%
  	}
  	\toprule
  	WS & +10 & +20 & +30 & +40 & +50 & +60 & +70 \\
  	\midrule
        \midrule
  	25 & \textbf{0.204} & 0.275 & 0.318 & 0.344 & 0.375 & 0.386 & 0.405\\
        35 & \textbf{0.216} & 0.290 & 0.322 & 0.357 & 0.372 & 0.394 & --\\
  	45 & \textbf{0.215} & 0.287 & 0.332 & 0.357 & 0.380 & --    & --\\
        55 & \textbf{0.202} & 0.283 & 0.327 & 0.357 & --    & --    & --\\
        65 & \textbf{0.209} & 0.286 & 0.330 & --    & --    & --    & --\\
        75 & \textbf{0.212} & 0.291 & --    & --    & --    & --    & --\\
        85 & \textbf{0.215} & --    & --    & --    & --    & --    & --\\
  	\bottomrule
  \end{tabu}%
\end{table}

We show the quantitative results of the forecasting trajectories in Table~\ref{tab:fore_mes} using the Mean Squared Error (MSE) between the ground truth trajectories and the forecasted trajectories as the evaluation metric. In the table, we use `WS' to denote the window size, and `+x' to represent the length of forecasted sequence. For instance, WS of 25 and x of 20 represent and input sequence consisting of 25 timestamps and forecasting future 20 timestamps. From Table~\ref{tab:fore_mes}, we see a distinct trend where the MSE is directly proportional to the length of the forecasting sequence for a fixed window size, i.e., as the length of the forecasting sequence increases, the MSE also increases. However, when forecasted sequences of the same length, we observe a weak linear trend between the window size and the MSE scores in Table~\ref{tab:fore_mes}, in other words, the MSE slightly increases as the window size increases, which suggests that smaller input windows are more likely to result in more precise forecasting when generating a fixed-length sequence.

\begin{figure}[t!]
    \centering
    \includegraphics[width=\linewidth]{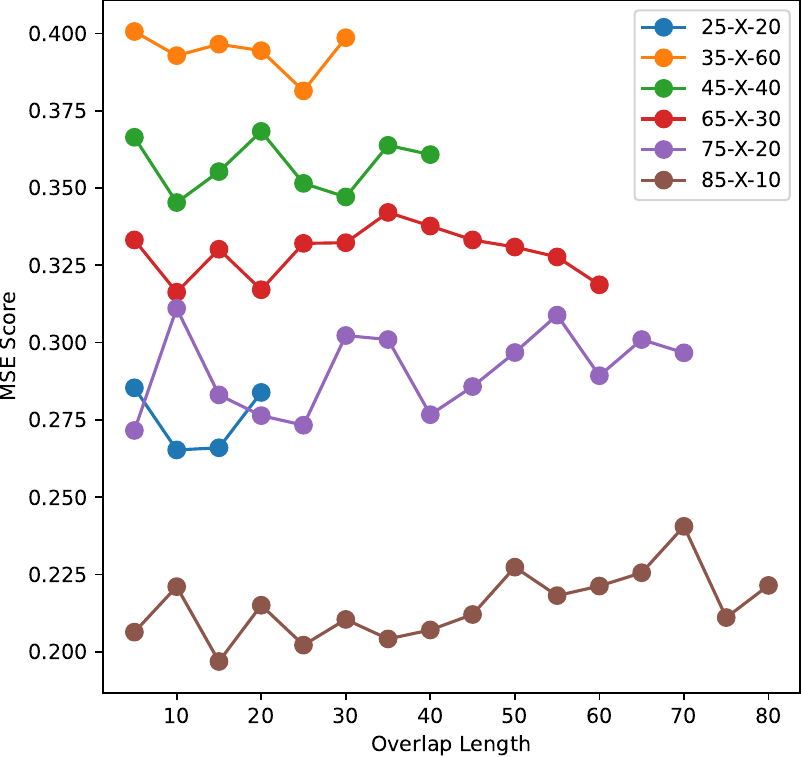}
    \caption{MSE scores of 6 fixed-length pairs of input and forecasted sequences with varying overlap length. All pairs (dotted lines) share the same $x$ and $y$ axis. Input window sizes are $25$, $35$, $45$, $65$, $75$, and $85$, and forecasting lengths are $20$, $60$, $40$, $30$, $20$, and $10$, corresponding to each line. Lengths of overlap range from 5 to $20$, $30$, $40$, $60$, $70$, and $80$, respectively, with the same step size of $5$.}
    \label{fig:overlap_compare}
\end{figure}

We conduct multiple trials by varying $l_{overlap}$, and see no evidence that the length of overlap data affects the accuracy of forecasting output trajectory. Figure~\ref{fig:overlap_compare} shows experimental results of $l_{window}$ and $l_{forecast}$, where $l_{window}$ takes on the values $25$, $35$, $45$, $65$, $75$, and $85$, and $l_{forecast}$ takes on the values $20$, $60$, $40$, $30$, $20$, and $10$, respectively corresponding to each lines in Figure~\ref{fig:overlap_compare}. For each pair of fixed $l_{window}$ and $l_{forecast}$ (each line in the figure), $l_{overlap}$ varies from $5$ to $l_{window}-5$ with stride $5$. We do not observe any trend indicating that $l_{overlap}$ significantly affects the forecasting accuracy in terms of MSE, As a result, we use the median of $l_{overlap}$ for each pair of $l_{window}$ and $l_{forecast}$ across the entire experiment.

\subsection{Authentication After Forecasting Results}
\setlength{\tabcolsep}{5.1pt}
\begin{table}[t!]
  \caption{EER of FCN as a Classifier with Forecasted Trajectory. `+x' means the length of forecasted sequence is x. `+0' means with no forecasting}
  \label{tab:exp3fcn}
  \footnotesize
  \centering%
  \begin{tabu}{
             c| *{8}{c}
            *{9}{c}%
  	}
  	\toprule
  	WS & +0 & +10 & +20 & +30 & +40 & +50 & +60 & +70 \\
  	\midrule
        \midrule
  	25 & 0.121 & 0.099 & 0.093 & 0.089 & 0.084 & \textbf{0.082} & 0.086 & 0.083\\
        35 & 0.101 & 0.085 & 0.082 & 0.077 & 0.072 & \textbf{0.067} & 0.073 & --\\
  	45 & 0.082 & 0.079 & 0.070 & 0.069 & \textbf{0.061} & 0.063 & --    & --\\
        55 & 0.082 & 0.068 & 0.063 & 0.057 & \textbf{0.055} & --    & --    & --\\
        65 & 0.075 & 0.063 & 0.058 & \textbf{0.052} & --    & --    & --    & --\\
        75 & 0.062 & 0.060 & \textbf{0.059} & --    & --    & --    & --    & --\\
        85 & 0.071 & \textbf{0.066}  & --    & --    & --    & --    & --    & --\\
  	\bottomrule
  \end{tabu}%
\end{table}

\begin{table}[t!]
  \caption{EER of Transformer Encoder as a Classifier with Forecasted Trajectory. `+x' means the length of forecasted sequence is x. `+0' means with no forecasting}
  \label{tab:exp3tf}
  \footnotesize%
  \centering%
  \begin{tabu}{%
            c| *{8}{c}
  	  	*{9}{c}%
  	}
  	\toprule
  	WS & +0 & +10 & +20 & +30 & +40 & +50 & +60 & +70 \\
  	\midrule
        \midrule
  	25 & 0.115 & 0.097 & 0.091 & 0.086 & \textbf{0.080} & 0.081 & 0.081 & 0.084\\
        35 & 0.097 & 0.080 & 0.075 & 0.070 & 0.068 & \textbf{0.064} & 0.065 & --\\
  	45 & 0.083 & 0.069 & 0.064 & 0.061 & 0.054 & \textbf{0.053} & --    & --\\
        55 & 0.072 & 0.062 & 0.057 & 0.054 & \textbf{0.049} & --    & --    & --\\
        65 & 0.064 & 0.057 & 0.053 & \textbf{0.048} & --    & --    & --    & --\\
        75 & 0.057 & 0.055 & \textbf{0.051} & --    & --    & --    & --    & --\\
        85 & 0.064 & \textbf{0.055} & --    & --    & --    & --    & --    & --\\
  	\bottomrule
  \end{tabu}%
\end{table}

In Table~\ref{tab:exp3fcn} and Table~\ref{tab:exp3tf}, we summarize the results using EER, where `WS' and `+x' are the same as those in Table~\ref{tab:fore_mes} and stand for the window size and the length of forecasted sequence. We use `+0' to represent no forecasting, i.e., the EER scores in the `+0' column are directly from Table~\ref{tab:exp1}. We compare the authentication performance between models with and without forecasting by calculating the EER reduction. We compute the EER reduction by subtracting the lowest EER score obtained from the results with forecasting sequences from the without forecasting EER score, then we divide the difference by the without forecasting EER score, giving us a percentage that represents the degree on improve authentication performance. We observe from Tables~\ref{tab:exp3fcn} and \ref{tab:exp3tf} that authentication using the forecasted trajectory outperforms that without forecasting for all window sizes. The lowest EER scores appear in columns for forecasted sequences as opposed to the first column without forecasting.
Results show that without forecasting, EER is higher, ranging over 0.062-0.121 and 0.055-0.115 respectively for the FCN and Transformer over the various WS values. Overall, the Transformer model provides lower EER values. With forecasting, we see consistent reduction in EER values. Lowest EERs for FCN and Transformer are 0.052 and 0.048, respectively both at WS of 65 and +x of +30. Reduction is higher for smaller WS, as more data about the user behavior can be forecasted, with a maximum drop of 0.035 from 0.115 to 0.080 (WS = 25, +x = +40) for the Transformer and a maximum drop of 0.039 from 0.121 to 0.082 (WS = 25, +x = +50) for the FCN. These drops suggest that our approach of forecasting future behavior improves authentication over not using forecasting.

For a test input sample from the user prior to forecasting, we obtain forecasting and authentication times of 3.50-4.28 milliseconds using the FCN and 4.33-4.99 milliseconds using the Transformer, i.e., $<$5 milliseconds. Given that the 135 timestamps span 3 seconds of data, timestamps are separated by 22.22 milliseconds. Forecasting and authentication, even for +70 or 1.55 seconds into the future, occurs well before data at the next timestamp is acquired. In theory, even if an attacker tried to break the system after a single timestamp of acquiring the first WS timestamps, our system can forecast and show higher-assurance authentication before the attacker can break the system. In practice, as our results show for the non-forecasted case, the attacker will require several more timestamps of data for higher assurance. For instance, to acquire an EER of around 0.057 an attacker using a classifier such as our Transformer will need the user to have provided 75 timestamps or 1.67 seconds worth of data according to Table~\ref{tab:exp3tf}. We can acquire a lower EER, with just 45 timestamps of data or 1 second of data by conducting forecasting to +40 or +50 timestamps, and the forecasting occurs within 5 milliseconds, i.e., by 1.005 seconds we will have gotten ahead of the attacker for an authentication system that operates with an EER of 0.057. Our approach thus enables early authentication to circumvent an attacker, enabling more secure systems.

\section{Discussion}
In this paper, we present the first approach that uses motion forecasting for behavioral biometrics in VR. We use a Transformer-based model to forecast motion trajectories given an initial trajectory of a user performing an action in VR. We merge the initial and forecasted trajectory and perform authentication. We compare the performance of two classifiers, a Transformer encoder and FCN, and demonstrate the effectiveness of our approach using the 41-subject ball-throwing dataset of Miller et al.~\cite{miller2022combining,miller2022temporal}. We show that our approach of forecasting provides a lower EER of 0.053 with 45 timestamps worth of data, as compared to an authentication without forecasting, where the lowest EER is 0.057 with a longer sample of data needed. Forecasting and authentication is performed within 5 milliseconds, i.e., within less than a single timestamp and well within the amount of time needed by an attacker to snoop the amount of user information to acquire the same level of authentication success. 

An important issue is that, though our method circumvents an attacker snooping the user-provided motion, it now enables an in-person attacker performing mimicry of a user's motion using the VR system to attack the system by providing a lower quantity of mimicked data, a task that may be easier for the attacker than precisely mimicking the full range of the user's data. A potential approach to circumvent this may be to design a version of a 2-factor authentication system, where the 2nd factor is the complete user trajectory, and the forecasted trajectory is compared to the complete trajectory, which is likely to be less precise for the attacker.

Our approach uses a ball-throwing task, which has a starting point, i.e. lifting the ball, and an end goal, i.e. attempting to hit the target, with little variability in the intermediary steps. Critical VR applications may have intermediary steps with high variability within and across users. For example, in a banking application we can have different intermediary steps between the starting point, i.e. the user opening the door, to the ending goal, i.e. depositing a check. In one session after opening the door and before depositing the check a user may speak to a teller or in another session look at the newest bank rates. These differences in intermediary steps may vary for the same user between sessions, for example, a user looking at the new bank rates at the start of a month. The intermediary steps may also vary between users, where one user may always speak to a teller before depositing a check while another user directly deposits the check. While it may seem that the variable intermediary steps can make motion forecasting challenging, they are no different from the unpredictable behavior of pedestrians in autonomous driving~\cite{zeng2021lanercnn,zhou2022hivt,huang2022multi,kong2022human,liu2021multimodal,yuan2021agentformer}. In future, we will investigate the robustness of Transformer-based forecasting models in complex VR scenarios with multiple intermediary pathways, such as a person depositing a check in a bank or a student taking an examination. We also plan to investigate motion forecasting for authentication using datasets such as the Alyx dataset, released mid November 2023, that contain more diverse behavior~\cite{rack2023alyx}.

\bibliographystyle{IEEEtran}
\bibliography{template}

\end{document}